\documentclass{article}
\usepackage{spconf,amsmath,graphicx,cite}

\newcommand{\bm}[1]{{\mbox{\boldmath $#1$}}}

\title{HIERARCHICAL KNOWLEDGE DISTILLATION FOR DIALOGUE SEQUENCE LABELING}
\name{\begin{tabular}{c}Shota Orihashi, Yoshihiro Yamazaki, Naoki Makishima, Mana Ihori, \\
Akihiko Takashima, Tomohiro Tanaka, Ryo Masumura\end{tabular}}
\address{NTT Media Intelligence Laboratories, NTT Corporation, Japan}
\begin{document}
%
\maketitle
\begin{abstract}
This paper presents a novel knowledge distillation method for dialogue sequence labeling. Dialogue sequence labeling is a supervised learning task that estimates labels for each utterance in the target dialogue document, and is useful for many applications such as dialogue act estimation. Accurate labeling is often realized by a hierarchically-structured large model consisting of utterance-level and dialogue-level networks that capture the contexts within an utterance and between utterances, respectively. However, due to its large model size, such a model cannot be deployed on resource-constrained devices. To overcome this difficulty, we focus on knowledge distillation which trains a small model by distilling the knowledge of a large and high performance teacher model. Our key idea is to distill the knowledge while keeping the complex contexts captured by the teacher model. To this end, the proposed method, hierarchical knowledge distillation, trains the small model by distilling not only the probability distribution of the label classification, but also the knowledge of utterance-level and dialogue-level contexts trained in the teacher model by training the model to mimic the teacher model's output in each level. Experiments on dialogue act estimation and call scene segmentation demonstrate the effectiveness of the proposed method.
\end{abstract}
\section{Introduction}
\label{sec:introduction}

With the progress of automatic speech recognition technologies, expectations for the understanding and utilization of linguistic information present in human-to-human dialogs are increasing. For example, by understanding contact center telephone dialogue documents, a service for discovering customer needs and issues with the center has been developed \cite{Mamou06, Byrd08, Higashinaka10, Tamura11, Chastagnol11, Ando17}.

In this paper, we focus on utterance-level dialogue sequence labeling, a key component in dialogue document understanding. Dialogue sequence labeling is often modeled as a supervised learning task that estimates labels for each utterance when given a dialogue document; it is useful in many applications such as topic segmentation \cite{Yu16, Tsunoo17a, Tsunoo17b}, dialogue act estimation \cite{Tran17, Kumar18, Chen18, Jiao19, Raheja19, Yu19}, and call scene segmentation \cite{Masumura18, Orihashi20, Masumura21}. To understand dialogue documents, it is necessary to consider who spoke what and in what order. Therefore, these techniques often adopt a hierarchically-structured model consisting of an utterance-level network and a dialogue-level network to capture contexts not only within an utterance but also between utterances \cite{Masumura18}. In addition, an effective self-supervised pretraining method using only unlabeled data has been proposed \cite{Masumura21}.

Capturing dialogue documents precisely demands a large number of parameters for both the utterance-level and the dialogue-level networks. However, label inference using such large models requires a rich computation environment. Unfortunately, it is difficult to prepare such an environment, especially when we need to process multiple inferences in parallel or process inference on a device with low computing power such as a mobile device. Therefore, using a large model hinders the adoption of various services.

To overcome the difficulties created by using large models, we focus on knowledge distillation; a small student model with just a few parameters is trained by distilling the knowledge of a large and high performance teacher model so replicate the teacher's performance \cite{Bucilua06, Hinton14, Romero15}. In recent years, knowledge distillation techniques have been successful in the natural language processing field; examples include neural machine translation \cite{Kim16, Freitag17, Tan19} and compression of bidirectional encoder representations from Transformers (BERT) \cite{Vaswani17, Devlin18, Sun19, Sanh19, Jiao20, Sun20}. The strength of knowledge distillation is that the student model can be trained to mimic the behavior of the teacher model. For dialogue sequence labeling, it is especially important to mimic the behavior of the teacher model faithfully to keep full performance while reducing the model size because dialogue sequence labeling is a task in which complex contexts at the utterance-level and the dialogue-level must be precisely captured. Knowledge distillation is seen as potentially able to overcome the difficulty of using a large model for dialogue sequence labeling, but no truly effective knowledge distillation technique for dialogue sequence labeling has been described so far.

In this paper, we propose a novel knowledge distillation method for dialogue sequence labeling. Our key idea is to train a small student model by distilling the knowledge of utterance-level and dialogue-level contexts while retaining the complex contexts captured by the large teacher model. To this end, our method, hierarchical knowledge distillation, not only trains the student model so that the probability distribution of its output labels approaches that of the teacher model, but also trains the student model so that the outputs of the utterance-level and the dialogue-level networks of the student model approach those of the teacher model. By distilling the knowledge of complex contexts from the large teacher model via hierarchical knowledge distillation, our method enables us to train the small student model without losing the ability to capture contexts within an utterance and between utterances as captured by the teacher model. Our experiments on dialogue act estimation using the switchboard dialogue act (SwDA) corpus \cite{Jurafsky97, Stolcke00} and call scene segmentation using a simulated Japanese contact center dialogue dataset demonstrate the effectiveness of the proposed method.

Our contributions are summarized as follows:
\begin{itemize}
\item We provide an effective knowledge distillation method for dialogue sequence labeling that distills not only the probability distribution of the label classification \cite{Hinton14}, but also the knowledge of utterance-level and dialogue-level contexts. To the best of our knowledge, this is the first method to achieve knowledge distillation for dialogue sequence labeling.
\item We conduct ablation experiments on dialogue act estimation and call scene segmentation tasks that analyze the effectiveness of the proposed method. We also provide the results achieved by combining self-supervised pretraining \cite{Masumura21} and the proposed method. 
\end{itemize}

\section{Related work}
\label{sec:related}

\subsection{Utterance-level dialogue sequence labeling}
\label{ssec:labeling}

Utterance-level dialogue sequence labeling is being used for topic segmentation \cite{Yu16, Tsunoo17a, Tsunoo17b}, dialogue act estimation \cite{Tran17, Kumar18, Chen18, Jiao19, Raheja19, Yu19}, and call scene segmentation \cite{Masumura18, Orihashi20, Masumura21}. Hierarchically structured models consisting of utterance-level and dialogue-level neural networks are often used to efficiently capture contexts within an utterance and between utterances, and an effective self-supervised pretraining method has been proposed \cite{Masumura21}. If a hierarchical model is used for dialogue sequence labeling, a large number of parameters are needed to train a model that offers high accuracy. In this paper, to train a highly accurate model that has just a few parameters, we introduce a knowledge distillation technique to utterance-level dialogue sequence labeling.

\subsection{Knowledge distillation}
\label{ssec:distillation}

Knowledge distillation is a technique to train a small student model efficiently by utilizing the knowledge of a large and high performance teacher model without significant performance loss \cite{Bucilua06}. One of the early methods trains the student model so that the probability distribution of the output label of the student model approaches that of the teacher model by utilizing soft target loss \cite{Hinton14}. Another method trains the student model so that the hidden layers' outputs of the student model approach those of the teacher model \cite{Romero15}. Successful knowledge distillation techniques have recently been reported in the natural language processing field \cite{Kim16, Freitag17, Tan19, Sun19, Sanh19, Jiao20, Sun20}. In this paper, we propose a knowledge distillation method for dialogue sequence labeling. To retain the ability to capture contexts within an utterance and between utterances, we train the student model so that the outputs of the utterance-level and the dialogue-level networks of the student model approach those of the teacher model.

\section{Utterance-level dialogue sequence labeling}
\label{sec:labeling}

This section describes utterance-level dialogue sequence labeling in dialogue documents. This task estimates utterance-level label sequence $\bm{Y}=\{y_1,\cdots,y_T\}$ from input utterance sequence $\bm{X}=\{x_1,\cdots,x_T\}$ using neural networks, where the $t$-th utterance, $x_t$, consists of token sequence $\{w_{t,1},\cdots,w_{t,{K_t}}\}$; $K_t$ is number of tokens in the $t$-th utterance. The $t$-th label, $y_t$, is an element of $\mathcal{Y}$, where $\mathcal{Y}$ is the set of labels. Label types are task dependent, for example dialogue act labels for dialogue act estimation and call scene labels for call scene segmentation.

In our dialogue sequence labeling, $y_t$ is estimated from $\{x_1,\cdots,x_t\}$ in an online manner. For this, conditional probability $P(y_t \mid x_1,\cdots,x_t,\bm{\Theta})$ is modeled, where $\bm{\Theta}$ represents a model parameter. The $t$-th label can be categorized by:
\newcommand{\argmax}{\mathop{\rm arg~max}\limits}
\begin{equation}
\hat{y}_t = \argmax_{y_t \in \mathcal{Y}} P(y_t \mid x_1,\cdots,x_t,\bm{\Theta}).
\end{equation}
In this paper, we assume that $P(y_t \mid x_1,\cdots,x_t,\bm{\Theta})$ is modeled by the Transformer encoder \cite{Vaswani17} and hierarchical long short-term memory recurrent neural networks (LSTM-RNNs). Figure 1 shows the structure of the labeling model.

\begin{figure*}[t]
\centering
\centerline{\includegraphics[width=0.7\linewidth]{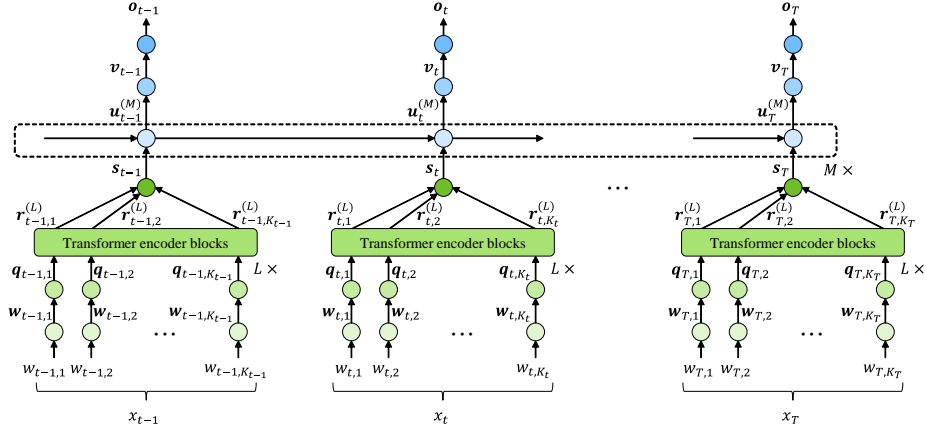}}
\caption{Structure of dialogue sequence labeling model.}
\label{fig:model}
\end{figure*}

In the utterance-level network, each token is first converted into a continuous vector representation. The continuous vector representation of the $k$-th token in the $t$-th utterance is given by:
\begin{equation}
\bm{w}_{t,k}={\rm Embedding}(w_{t,k};\bm{\theta^w}),
\end{equation}
where ${\rm Embedding}()$ is a linear transformational function that embeds a symbol into a continuous vector, and $\bm{\theta^w}$ is the trainable parameter. Continuous vectors $\bm{w}_{t,k}$ are then converted into $\bm{q}_{t,k}$ for input to the Transformer encoder block as:
\begin{equation}
\bm{q}_{t,k}={\rm AddPosEnc}(\bm{w}_{t,k}),
\end{equation}
where ${\rm AddPosEnc}()$ is a function that adds a continuous vector in which position information is embedded. The Transformer encoder forms hidden representations $\bm{R}_t^{(L)}=\{\bm{r}_{t,1}^{(L)},\cdots,\bm{r}_{t,{K_t}}^{(L)}\}$ from $\bm{Q}_t=\{\bm{q}_{t,1},\cdots,\bm{q}_{t,{K_t}}\}$ by using $L$ Transformer encoder blocks. The $l$-th Transformer encoder block forms the $l$-th hidden representations $\bm{R}_t^{(l)}$ from the lower layer outputs $\bm{R}_t^{(l-1)}$ as:
\begin{equation}
\bm{R}_t^{(l)}={\rm TransformerEnc}(\bm{R}_t^{(l-1)};\bm{\theta^r}),
\end{equation}
where $\bm{R}_t^{(0)}=\bm{Q}_t$, and ${\rm TransformerEnc}()$ is a Transformer encoder block that consists of a scaled dot product multi-head self-attention layer and a position-wise feed-forward network \cite{Vaswani17}. $\bm{\theta^r}$ represents the trainable parameter. The hidden representations are then summarized as an utterance representation by a self-attention mechanism \cite{Chang19}. The $t$-th utterance continuous representation is calculated as:
\begin{equation}
\bm{s}_t={\rm SelfAttention}(\bm{r}_{t,1}^{(L)},\cdots,\bm{r}_{t,{K_t}}^{(L)};\bm{\theta^s}),
\end{equation}
where ${\rm SelfAttention}()$ is a transformational function that converts into a fixed-length vector by the self-attention mechanism; $\bm{\theta^s}$ is the trainable parameter.

In the dialogue-level network, interaction information from start-of-dialogue to the $t$-th utterance is incrementally embedded into a continuous vector representation. The $t$-th continuous vector representation that embeds all dialogue context sequential information up to the $t$-th utterance $\bm{u}_t^{(M)}$ is calculated from $\{\bm{s}_1,\cdots,\bm{s}_t\}$ by using $M$ LSTM-RNN layers. The $m$-th LSTM-RNN layer forms the $m$-th hidden representations $\bm{u}_t^{(m)}$ from the lower layer outputs $\{\bm{u}_1^{(m-1)},\cdots,\bm{u}_t^{(m-1)}\}$ as:
\begin{equation}
\bm{u}_t^{(m)}={\rm LSTM}(\bm{u}_1^{(m-1)},\cdots,\bm{u}_t^{(m-1)};\bm{\theta^u}),
\end{equation}
where $\bm{u}_t^{(0)}=\bm{s}_t$, ${\rm LSTM}()$ is a function of the unidirectional LSTM-RNN layer, and $\bm{\theta^u}$ represents the trainable parameter.

In the output layer, predictive probabilities of the labels for the $t$-th utterance $\bm{o}_t$ are defined using logits $\bm{v}_t=[v_{t,1},\cdots,v_{t,|\mathcal{Y}|}]$ as:
\begin{equation}
\bm{v}_t={\rm FeedForward}(\bm{u}_t^{(M)};\bm{\theta^v}),
\end{equation}
\begin{equation}
\bm{o}_t={\rm Softmax}(\bm{v}_t),
\end{equation}
where ${\rm FeedForward}()$ is a function of a fully-connected feed forward neural network, $\bm{\theta^v}$ is a trainable parameter, and ${\rm Softmax}()$ is a softmax function. $\bm{o}_t$ corresponds to $P(y_t \mid x_1,\cdots,x_t,\bm{\Theta})$.

The model parameters $\bm{\Theta}=\{\bm{\theta^w},\bm{\theta^r},\bm{\theta^s},\bm{\theta^u},\bm{\theta^v}\}$ can be optimized by preparing training dataset $\mathcal{D}=\{(\bm{X}^1,\overline{\bm{Y}}^1),\cdots,\\(\bm{X}^N,\overline{\bm{Y}}^N)\}$, where $\bm{X}^n$ and $\overline{\bm{Y}}^n$ are input utterance sequence and reference utterance-level label sequence in the $n$-th dialogue, respectively. In this case, cross-entropy loss, named hard target loss, is computed by:
\begin{equation}
\mathcal{L}_{\rm HT}=-\frac{1}{N}\sum_{n=1}^N\left(\frac{1}{T_n}\sum_{t=1}^{T_n}\sum_{y\in\mathcal{Y}}\overline{o}_{t,y}^n\log{o_{t,y}^n}\right),
\end{equation}
where $\overline{\bm{o}}_t^n=[\overline{o}_{t,1}^n,\cdots,\overline{o}_{t,|\mathcal{Y}|}^n]$ and $\bm{o}_t^n=[o_{t,1}^n,\cdots,o_{t,|\mathcal{Y}|}^n]$ are the reference and estimated probabilities of label $y$ for the $t$-th utterance in the $n$-th dialogue, respectively. $T_n$ is the number of utterances in the $n$-th dialogue. Note that $\overline{\bm{o}}_t^n$ is a one-hot vector.

When self-supervised pretraining \cite{Masumura21} is utilized, parameters $\{\bm{\theta^w},\bm{\theta^r},\bm{\theta^s},\bm{\theta^u}\}$ are initialized by pretraining using unlabeled data, and then parameters $\bm{\Theta}$ are optimized with $\mathcal{L}_{\rm HT}$ in the same way as above.

\section{Proposed method}
\label{sec:method}

This section details our proposed knowledge distillation method for utterance-level dialogue sequence labeling. The main idea of the proposed method, hierarchical knowledge distillation, is to train a small student model by distilling the knowledge of utterance-level and dialogue-level complex contexts captured by a large teacher model. To this end, our method trains the student model so that the probability distribution of the output labels approaches that of the large teacher model by utilizing soft target loss \cite{Hinton14}. Not only that, our method trains the student model so that the outputs of the utterance-level and the dialogue-level networks of the student model approach those of the teacher model. By distilling the knowledge to capture complex contexts trained in the teacher model, our method efficiently trains the small student model so that it offers high accuracy.

Figure 2 outlines the proposed method. Our hierarchical knowledge distillation proposal trains the student model by distilling the knowledge of the teacher model by optimizing the student model using a loss function that is a combination of four components: hard target loss (9), soft target loss, utterance-level context loss, and dialogue-level context loss. 

\subsection{Soft target loss}
\label{ssec:soft}

Soft target loss aims to bring the student model's probability distribution of the output labels closer to that of the teacher model \cite{Hinton14}. To calculate soft target loss, the probability distribution is computed from logits $\bm{v}_t$ by:
\begin{equation}
\bm{z}_t={\rm SoftmaxWithTemperature}(\bm{v}_t;\tau),
\end{equation}
where ${\rm SoftmaxWithTemperature}()$ is a softmax function with temperature, and $\tau$ is a hyper-parameter that represents temperature \cite{Hinton14}. Soft target loss is thus defined as:
\begin{equation}
\mathcal{L}_{\rm ST}=-\frac{\tau^2}{N}\sum_{n=1}^N\left(\frac{1}{T_n}\sum_{t=1}^{T_n}\sum_{y\in\mathcal{Y}}\tilde{z}_{t,y}^n\log{z_{t,y}^n}\right),
\end{equation}
where $\tilde{\bm{z}}_t^n=[\tilde{z}_{t,1}^n,\cdots,\tilde{z}_{t,|\mathcal{Y}|}^n]$ and $\bm{z}_t^n=[z_{t,1}^n,\cdots,z_{t,|\mathcal{Y}|}^n]$ are the probabilities of label $y$ for the $t$-th utterance in the $n$-th dialogue estimated by the teacher model and the student model, respectively.

\subsection{Utterance-level context loss}
\label{ssec:utterance}

The proposed method aims to train the student model so that the utterance-level network of the student model mimics that of the teacher model. For this, utterance-level context loss is defined as the difference between the outputs of the utterance-level networks of the student and the teacher models. Utterance-level context loss is defined using mean squared error (MSE) as:
\begin{equation}
\mathcal{L}_{\rm UC}=\frac{1}{N}\sum_{n=1}^N\left(\frac{1}{T_n}\sum_{t=1}^{T_n}\|\tilde{\bm{s}}_t^n-\bm{s}_t^n\|_2^2\right),
\end{equation}
where $\tilde{\bm{s}}_t^n$ and $\bm{s}_t^n$ are the $t$-th utterance continuous vector representations of the teacher model and the student model, respectively. Note that the proposed method assumes that $\tilde{\bm{s}}_t^n$ and $\bm{s}_t^n$ have equal size. 

\begin{figure}[t]
\centering
\centerline{\includegraphics[width=\linewidth]{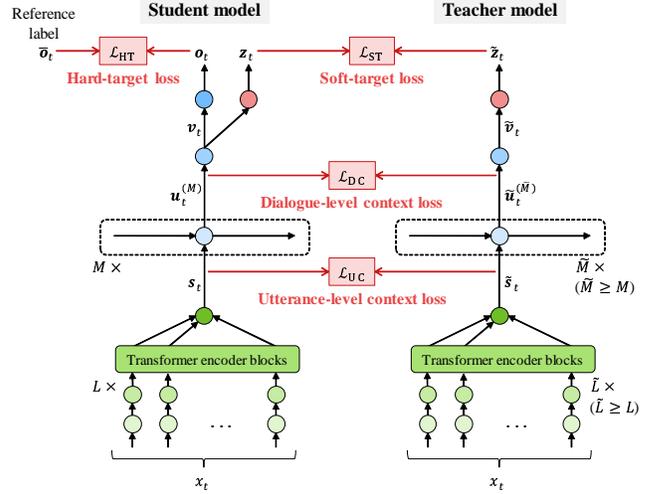}}
\caption{Outline of the proposed method.}
\label{fig:method}
\end{figure}

\subsection{Dialogue-level context loss}
\label{ssec:dialogue}

The proposed method also aims to train the student model so that the dialogue-level network of the student model mimics that of the teacher model. For this, dialogue-level context loss is defined as the difference between the outputs of the dialogue-level networks of the student and the teacher models. Dialogue-level context loss is defined using MSE as:
\begin{equation}
\mathcal{L}_{\rm DC}=\frac{1}{N}\sum_{n=1}^N\left(\frac{1}{T_n}\sum_{t=1}^{T_n}\|\tilde{\bm{u}}_t^{(\tilde{M}),n}-\bm{u}_t^{(M),n}\|_2^2\right),
\end{equation}
where $\tilde{\bm{u}}_t^{(\tilde{M}),n}$ and $\bm{u}_t^{(M),n}$ are the $t$-th continuous vector representations that embed all dialogue context sequential information up to the $t$-th utterance, of the teacher model and the student model, respectively. $\tilde{M}$ is the number of layers for the teacher model's dialogue-level network ($\tilde{M}\geq M$). Note that the proposed method assumes that $\tilde{\bm{u}}_t^{(\tilde{M}),n}$ and $\bm{u}_t^{(M),n}$ have equal size.

\subsection{Training}
\label{ssec:training}

For training, the parameters of the student model $\bm{\Theta}$ are optimized by using training dataset $\mathcal{D}$ with the application of combined loss. Combined loss is defined as:
\begin{equation}
\mathcal{L}=\mathcal{L}_{\rm HT}+\lambda\mathcal{L}_{\rm ST}+\alpha\mathcal{L}_{\rm UC}+\beta\mathcal{L}_{\rm DC},
\end{equation}
where $\lambda$, $\alpha$, and $\beta$ are hyper-parameters.

When self-supervised pretraining \cite{Masumura21} is utilized, parameters $\{\bm{\theta^w},\bm{\theta^r},\bm{\theta^s},\bm{\theta^u}\}$ are initialized in the pretraining using unlabeled data, and then parameters $\bm{\Theta}$ are optimized with $\mathcal{L}$ in the same way as above.

\begin{table}[t]
\caption{Details of the simulated Japanese contact center dialogue dataset.}
\label{tab:dataset}
\centering
\begin{tabular}{|l|rrr|}
\hline
\textbf{Business type} & \textbf{\#calls} & \textbf{\#utterances} & \textbf{\#tokens} \\
\hline\hline
Finance & 59 & 6,081 & 55,933 \\
Internet provider & 57 & 3,815 & 47,668 \\
Government unit & 73 & 5,617 & 48,998 \\
Mail-order & 56 & 4,938 & 46,574 \\
PC repair & 55 & 6,263 & 55,101 \\
Mobile phone & 61 & 5,738 & 51,061 \\
\hline
All & 361 & 32,452 & 305,351 \\
\hline
\end{tabular}
\end{table}

\section{Experiment}
\label{sec:experiment}

\subsection{Datasets}
\label{ssec:datasets}

We evaluated the proposed knowledge distillation method on two dialogue sequence labeling tasks: dialogue act estimation and call scene segmentation.

For dialogue act estimation, we used SwDA corpus \cite{Jurafsky97, Stolcke00}. SwDA corpus consists of 1,155 telephone calls between two people with no specific topic; it holds 205K utterances and 1.4M tokens. Each utterance is tagged with one dialogue act label, and each dialogue act label summarizes syntactic, semantic and pragmatic information about the corresponding utterance. SwDA corpus originally used over 200 kinds of dialogue act labels, but labels are usually clustered into 43 label-types such as {\it statement-non-opinion}, {\it acknowledge (backchannel)}, {\it statement-opinion}, and {\it agree/accept} \cite{Jurafsky97}. Following this, we used the clustered 43 dialogue act label-types. We split the SwDA corpus into 1,115 training dialogues and 19 test dialogues following the conventional approach \cite{Stolcke00}.

For call scene segmentation, we used a simulated Japanese contact center dialogue dataset consisting of 361 dialogues in six business fields. Details of the dataset are shown in Table 1. One dialogue means one telephone call between one operator and one customer; all utterances were manually transcribed. Each dialogue was divided into speech units using LSTM-RNN-based speech activity detection \cite{Eyben13} trained from various Japanese speech samples. We manually set five labels to define call scenes: {\it opening}, {\it requirement confirmation}, {\it response}, {\it customer confirmation}, and {\it closing} \cite{Masumura18}. We split the dataset into 324 training dialogues and 37 test dialogues. Only for call scene segmentation, we additionally prepared 4,000 unlabeled dialogues collected from various Japanese contact centers, and an additional 500 million unlabeled Japanese sentences collected from the Web to utilize self-supervised pretraining \cite{Masumura21}.

\subsection{Setups}
\label{ssec:setups}

We first trained the teacher model from the labeled dataset. For dialogue act estimation, we trained the teacher model using only labeled dataset from scratch. For call scene segmentation, we trained the teacher model utilizing the self-supervised pretraining \cite{Masumura21} using unlabeled dialogues and unlabeled sentences before training by using a labeled dataset. To evaluate the proposed knowledge distillation, we constructed student models by the following two training procedures. In \textbf{Baseline}, training used only the labeled dataset from scratch. In \textbf{Knowledge distillation}, training used only the labeled dataset and acquired the knowledge of the teacher model by utilizing the knowledge distillation proposal. Only for call scene segmentation, we constructed additional student models by the following two training procedures. In \textbf{Pretraining}, we utilized the self-supervised pretraining as used by the teacher model, and then trained using the labeled dataset. In \textbf{Pretraining + Knowledge distillation}, we utilized the self-supervised pretraining as used by the teacher model, and then trained utilizing the knowledge distillation proposal using the labeled dataset.

\begin{table}[t]
\caption{Details of the models and sizes.}
\label{tab:models}
\centering
\begin{tabular}{|l|rrrr|}
\hline
\textbf{} & \textbf{$L$} & \textbf{$M$} & \textbf{\#units} & \textbf{\#parameters} \\
\hline\hline
Teacher & 8 & 2 & 2,048 & 13.11M \\
\hline
S1 & 1 & 1 & 256 & 2.47M \\
S2 & 2 & 2 & 512 & 3.65M \\
\hline
\end{tabular}
\end{table}

\begin{table}[t]
\caption{Results in terms of classification accuracy for dialogue act estimation (\%).}
\label{tab:results}
\centering
\scalebox{0.97}[1.0]{
\begin{tabular}{|ll|rr|}
\hline
\textbf{} & \textbf{} & \textbf{S1} & \textbf{S2} \\
\hline\hline
\multicolumn{2}{|l|}{Teacher (common to S1 and S2)} & 72.79 & 72.79 \\
\hline
\multicolumn{2}{|l|}{Student} & & \\
& Baseline & 71.43 & 71.75 \\
\cline{2-4}
& Knowledge distillation w/o $\mathcal{L}_{\rm UC}, \mathcal{L}_{\rm DC}$ & 72.40 & 72.44 \\
& Knowledge distillation w/o $\mathcal{L}_{\rm UC}$ & 72.54 & 72.66 \\
& Knowledge distillation w/o $\mathcal{L}_{\rm DC}$ & 72.60 & 72.67 \\
& Proposed knowledge distillation & \textbf{72.69} & \textbf{72.81} \\
\hline
\end{tabular}
}
\end{table}

\begin{table*}[t]
\caption{Results in terms of classification accuracy for call scene segmentation (\%).}
\label{tab:results}
\centering
\begin{tabular}{|ll|rr|}
\hline
\textbf{} & \textbf{} & ~~~~~~~~~~\textbf{S1} & ~~~~~~~~~~\textbf{S2} \\
\hline\hline
\multicolumn{2}{|l|}{Teacher (common to S1 and S2)} & 91.28 & 91.28 \\
\hline
Student~~~~~ & Baseline & 87.22 & 87.94 \\
& Pretraining & 89.05 & 89.46 \\
\cline{2-4}
& Knowledge distillation w/o $\mathcal{L}_{\rm UC}, \mathcal{L}_{\rm DC}$ & 87.54 & 88.23 \\
& Knowledge distillation w/o $\mathcal{L}_{\rm UC}$ & 88.83 & 88.86 \\
& Knowledge distillation w/o $\mathcal{L}_{\rm DC}$ & 88.99 & 89.20 \\
& Proposed knowledge distillation & 89.81 & 91.10 \\
\cline{2-4}
& Pretraining + Knowledge distillation w/o $\mathcal{L}_{\rm UC}, \mathcal{L}_{\rm DC}$~~~~~ & 88.65 & 88.82 \\
& Pretraining + Knowledge distillation w/o $\mathcal{L}_{\rm UC}$ & 89.42 & 89.62 \\
& Pretraining + Knowledge distillation w/o $\mathcal{L}_{\rm DC}$ & 89.49 & 89.86 \\
& Pretraining + Proposed knowledge distillation & \textbf{89.98} & \textbf{91.26} \\
\hline
\end{tabular}
\end{table*}

Our experiments examined student models of two sizes: S1 and S2. Details of the models and their size, together with the teacher model, are shown in Table 2. In Table 2, $L$ and $M$ are the number of layers for utterance-level network and the dialogue-level network, respectively. Also, ``\#units" represents the inner outputs in the position-wise feed forward networks for Transformer encoder blocks. For all models, we defined the token vector representation as a 256-dimensional vector, and unit size of LSTM-RNN was set to 256. For the Transformer encoder blocks, the dimensions of the output continuous representations were set to 256, and the number of heads in the multi-head attentions was set to 4. Note that the teacher model is common to S1 and S2.

For training, the mini-batch size was set to five dialogues. The optimizer was RAdam \cite{Liu20} with the default setting. For knowledge distillation, parameters $\tau$, $\lambda$, $\alpha$, and $\beta$ were set to 5.0, 0.1, 0.05, and 0.05, respectively. We constructed five models by varying the initial parameters, and evaluated their average classification accuracy. Note that a part of the training dialogues was used for early stopping.

\subsection{Results}
\label{ssec:results}

The resulting classification accuracy values for dialogue act estimation are shown in Table 3. In the table, line 1 shows ideal accuracy achieved by the teacher model. Line 2 shows results yielded by training the student models from scratch. The results show that there is a performance gap between line 1 and line 2; this is due to a reduction in the number of parameters, see Table 2. Lines 3--6 show the results of knowledge distillation. Line 3 shows the results yielded by using only hard and soft target losses without utterance-level and dialogue-level context losses, which follows a previous method \cite{Hinton14}. The results on line 3 show performance improvements over the baseline, but the improvements were limited. Lines 4 and 5 show the results yielded by applying the knowledge distillation proposal without utterance-level context loss or dialogue-level context loss, respectively. The results on lines 4 and 5 show that the utilization of utterance-level or dialogue-level context loss improved performance compared with line 3. Line 6 shows the results achieved by the knowledge distillation proposal; the proposed method attained the best performance. Especially for S2, the accuracy of the proposed method exceeds that of the teacher model.

The resulting classification accuracy values for call scene segmentation are shown in Table 4. In the table, line 1 shows ideal accuracy values achieved by the teacher model. Line 2 shows results yielded by training the student models from scratch, and line 3 shows results yielded by utilizing the self-supervised pretraining. The results show that there is a performance gap between line 1 and lines 2 and 3 due to parameter reduction. Lines 4--7 show the results of knowledge distillation without pretraining the student models. Line 4 shows that using only hard and soft target losses yielded poor knowledge distillation performance. Lines 5 and 6 show that applying the knowledge distillation proposal without utterance-level context loss or dialogue-level context loss yielded limited performance improvements. Line 7 shows that the knowledge distillation proposal exceeds the performance of the results on lines 4--6. In addition, lines 8--11 show the results yielded by applying knowledge distillation with pretraining. The results on lines 8--10 show that full knowledge distillation performance was not attained when only a part of the loss was used. Note that lines 8--10 demonstrate improved performance compared to lines 4--6 due to pretraining. Line 11 shows that applying the proposed knowledge distillation with pretraining yielded the best performance of all other methods examined. S2 allowed the proposed method to most closely approach the accuracy of the teacher model.

The performance improvements attained by the knowledge distillation proposal are considered to be due to the fact that the proposed method could train the student model without losing the ability of the teacher model to capture contexts within an utterance and between utterances. Our results show that the proposed knowledge distillation method is an effective way of improving performance in small student models.

\section{Conclusions}
\label{sec:conclusions}

This paper has proposed a novel knowledge distillation method, hierarchical knowledge distillation, for dialogue sequence labeling. The key advance of our method is to distill the knowledge of the utterance-level and dialogue-level contexts captured by a large teacher model. To this end, our method utilizes utterance-level and dialogue-level context losses so that the outputs of the utterance-level and the dialogue-level networks of the student model approach those of the teacher model. Experiments on dialogue act estimation and call scene segmentation tasks showed that our method allows small student models to achieve better performance and that combining utterance-level and dialogue-level context losses is an effective approach to knowledge distillation for dialogue sequence labeling.


\bibliographystyle{IEEEbib}
\bibliography{refs}

\end{document}